\pdfoutput=1

\documentclass[11pt]{article}

\usepackage[]{naacl2021}

\usepackage{times}
\usepackage{latexsym}

\usepackage[T1]{fontenc}

\usepackage[utf8]{inputenc}

\usepackage{microtype}

\usepackage{graphicx}
\usepackage{textcomp}
\usepackage{paralist}

\title{PanGEA: The Panoramic Graph Environment Annotation Toolkit}

\author{Alexander Ku\thanks{\enspace First two authors contributed equally.} \quad Peter Anderson\footnotemark[1] \quad Jordi Pont-Tuset \quad Jason Baldridge\\
Google Research\\
\texttt{\normalsize{\{alexku, pjand, jponttuset, jridge\}@google.com }}}

\newcommand{\todo}[1]{}
\renewcommand{\todo}[1]{{\color{red} {#1}}}

\begin{document}
\maketitle
\begin{abstract}
PanGEA, the Panoramic Graph Environment Annotation toolkit, is a lightweight toolkit for collecting speech and text annotations in photo-realistic 3D environments. PanGEA immerses annotators in a web-based simulation and allows them to move around easily as they speak and/or listen. It includes database and cloud storage integration, plus utilities for automatically aligning recorded speech with manual transcriptions and the virtual pose of the annotators. Out of the box, PanGEA supports two tasks -- collecting navigation instructions and navigation instruction following -- and it could be easily adapted for annotating walking tours, finding and labeling landmarks or objects, and similar tasks. We share best practices learned from using PanGEA in a 20,000 hour annotation effort to collect the Room-Across-Room dataset. We hope that our open-source annotation toolkit and insights will both expedite future data collection efforts and spur innovation on the kinds of grounded language tasks such environments can support.
\end{abstract}

\section{Introduction}

The release of high-quality 3D building and street captures ~\cite{Matterport3D,mirowski2019streetlearn,mehta2020retouchdown,xiazamirhe2018gibsonenv,replica19arxiv} has galvanized interest in developing embodied navigation agents that can operate in complex human environments. Based on these environments, annotations have been collected for a variety of tasks including navigating to a particular class of object (ObjectNav)~\cite{batra2020objectnav}, navigating from language instructions aka vision-and-language navigation (VLN)~\cite{anderson2018vision,chen2019touchdown,Qi_2020_CVPR,ku2020room}, and vision-and-dialog navigation~\cite{thomason2020vision,hahn2020way}. To date, most of these data collection efforts have required the development of custom annotation tools.

To expedite future data collection efforts, in this paper we introduce PanGEA, an open-sourced annotation toolkit designed for these settings.\footnote{\url{github.com/google-research/pangea}} Specifically, PanGEA assumes an environment represented by discrete navigation graphs connecting high-resolution 360\textdegree~panoramas, where each node represents a unique viewpoint in the environment and actions involve moving between these viewpoints. Examples of suitable environments include the indoor buildings from Matterport3D \cite{Matterport3D} (using the navigation graphs from \citet{anderson2018vision}) and the street-level environments from StreetLearn \cite{mirowski2019streetlearn}. 

\begin{figure*}
    \centering
    \includegraphics[trim=9cm 9cm 9cm 0,height=0.47\linewidth]{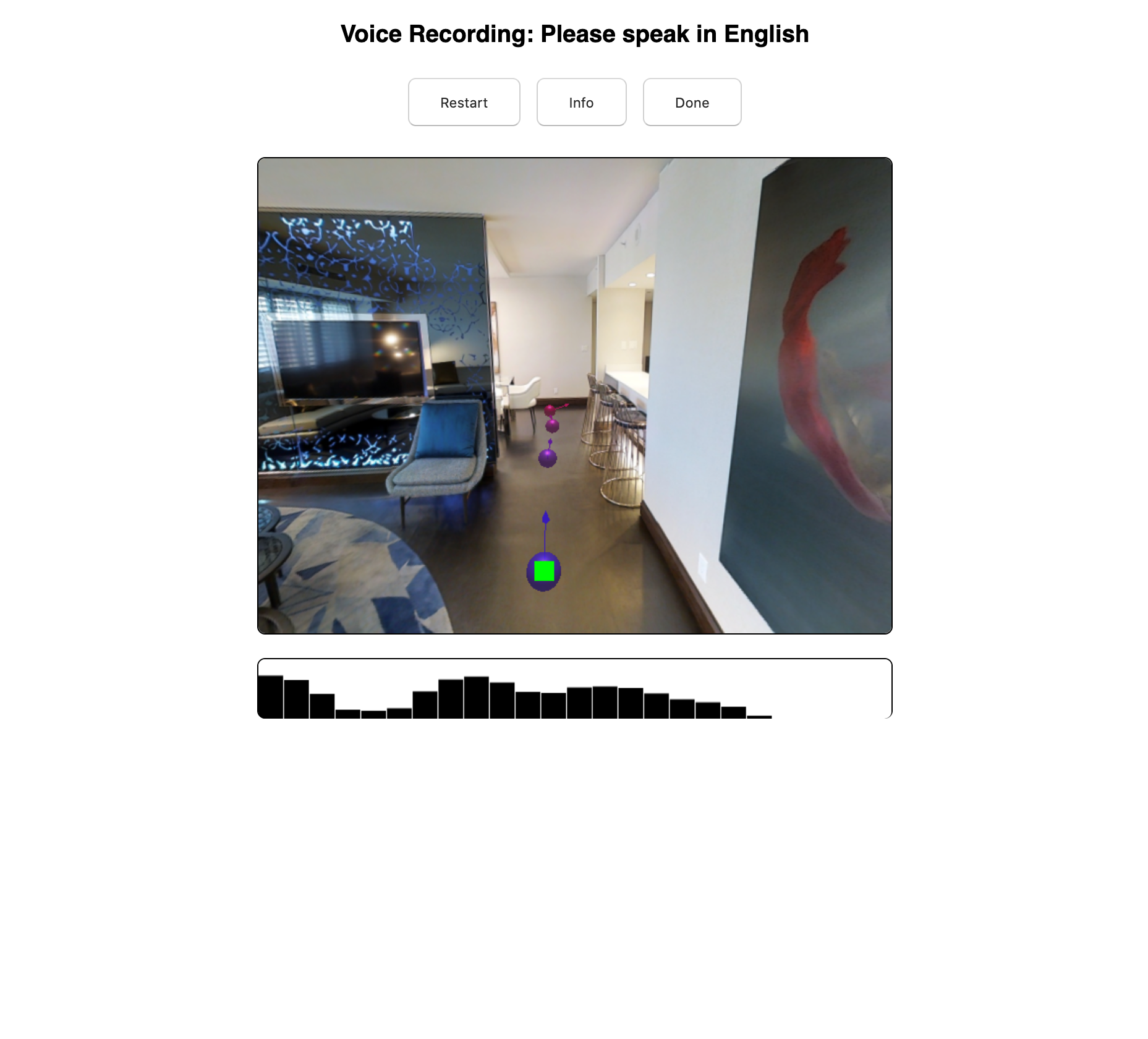}
    \includegraphics[trim=3cm 8cm 3cm 0,height=0.47\linewidth]{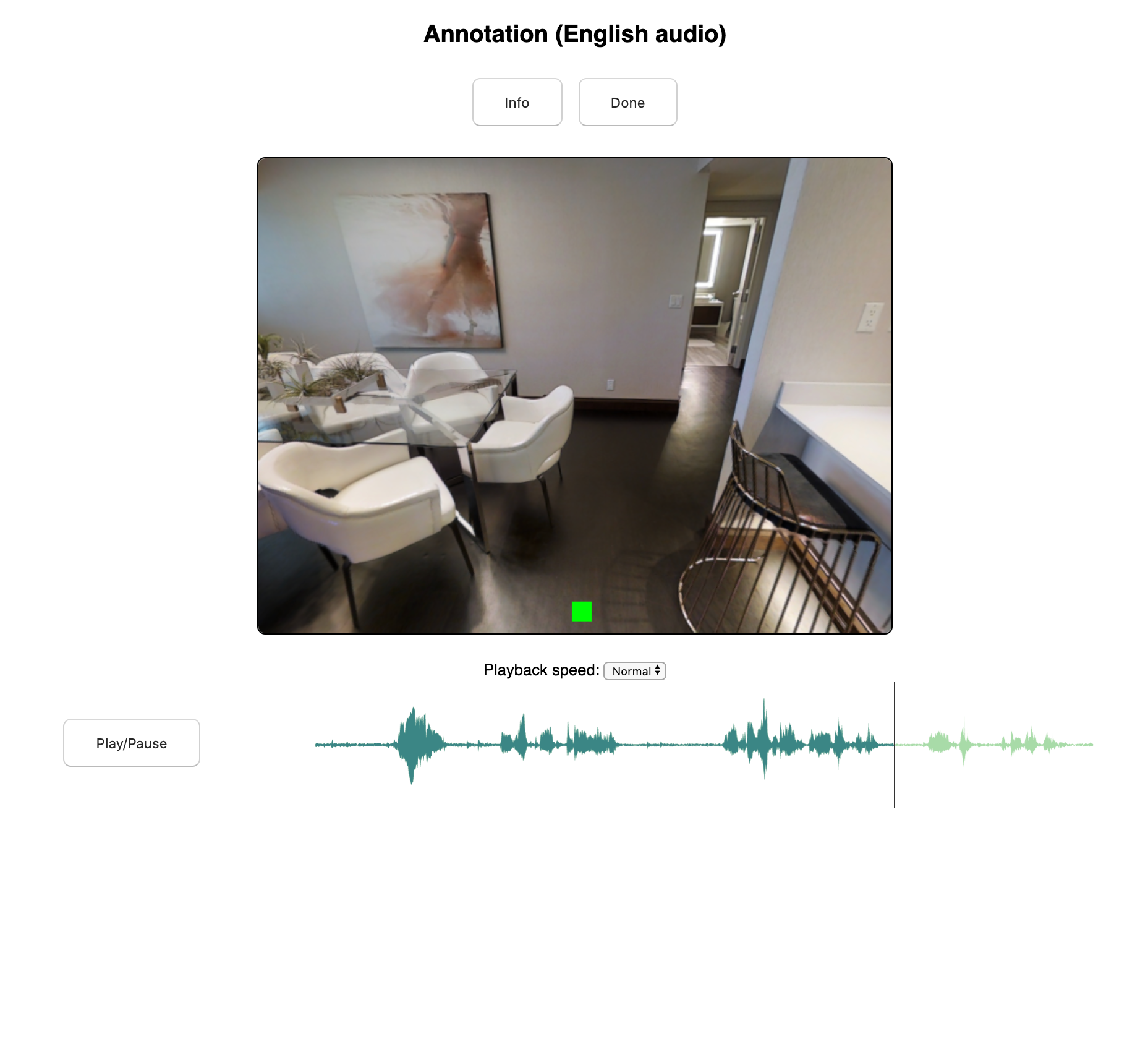}
    \caption{Screenshots of the PanGEA Guide and Follower interfaces. In the Guide task (left), Guides explore a given path while attempting to create a navigation instruction for others to follow. Guides can pause and restart the audio recording at any time. After recording is completed, Guides transcribe their own audio. In the Follower task (right), annotators listen to a Guide’s instructions and attempt to follow the intended path. Followers can skip around the Guide's audio using the audio waveform at bottom right. In both tasks, PanGEA tracks the annotators virtual camera pose and automatically aligns it with the Guide's audio transcript.}
    \label{fig:ui}
\end{figure*}

Out of the box, PanGEA supports two annotation modes: the \textit{Guide} task and the \textit{Follower} task. In the Guide task, Guides look around and move through an environment to follow a pre-defined path and attempt to create a navigation instruction for others to follow. In the Follower task, annotators listen to a Guide’s instructions and attempt to follow the path. These annotation modes are based on the Vision-and-Language Navigation (VLN) setting proposed by \citet{anderson2018vision}. However, compared to similar annotation tools, PanGEA includes substantial additional capabilities, notably: 

\begin{compactitem}
    \item annotation via voice recording (in addition to text entry)
    \item virtual pose tracking to record what annotators look at
    \item utilities for aligning a transcript of the words heard or uttered by each annotator with their visual perceptions and actions
    \item integration with cloud database and storage platforms
    \item a modular API facilitating easy extension to new tasks and new environments
\end{compactitem}

PanGEA has already been used in two papers. It was used to collect Room-Across-Room (RxR)~\cite{ku2020room}, a dataset of human-annotated navigation instructions in English, Hindi and Telugu which is the largest VLN dataset by an order of magnitude. PanGEA was also used to perform human evaluations of model-generated navigation instructions in \citet{discriminator21}. It could be trivially adapted to other tasks that combine annotation with movement, such as annotating walking tours, or finding and labeling particular landmarks or objects. 

We next describe PanGEA's capabitilities in more detail. In the final section we share some best practices learned from using PanGEA to collect RxR, which required more than 20,000 annotation hours.

\section{PanGEA Toolkit}

\begin{figure*}
    \centering
    \includegraphics[width=\linewidth]{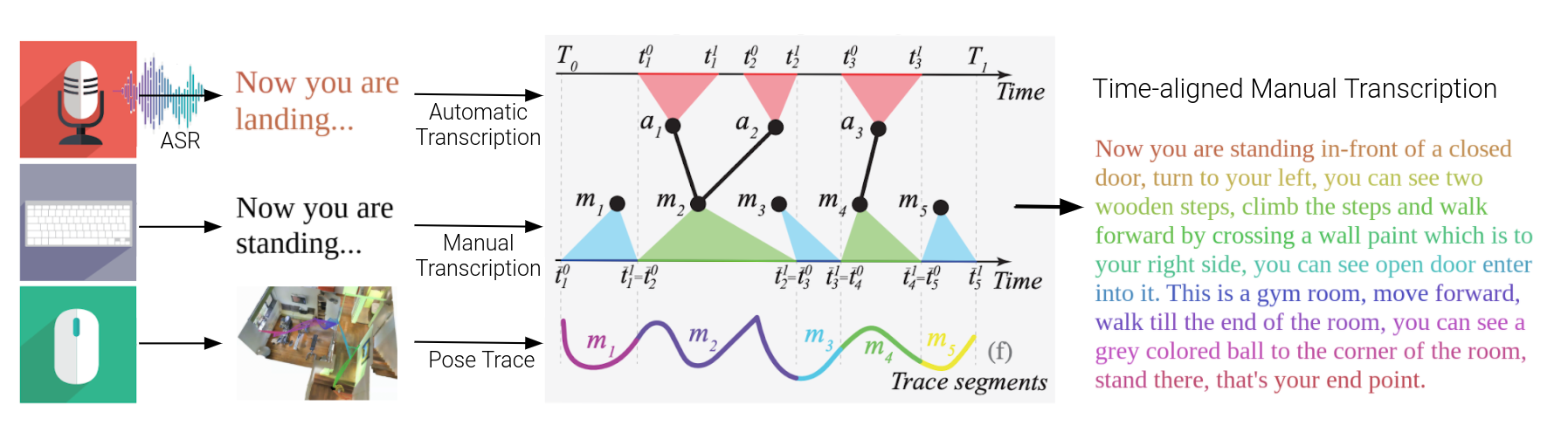}
    \caption{PanGEA time-aligns each annotators manual audio transcription (middle) to a \textit{pose trace} recording their virtual camera movements (bottom). This is achieved by first generating a noisy-but-timestamped automatic transcription (top), which is aligned with the manual transcription using dynamic time warping in order to propagate timestamps to the manual transcription. Figure adapted from \citet{PontTuset_eccv2020}}
    \label{fig:pose_trace}
\end{figure*}



\paragraph{Guide Task}
In the Guide task (Figure \ref{fig:ui}, left), Guides look around and move to explore an environment while recording an audio narration. For the RxR data collection, the Guide's movement was restricted to a particular path through the environment, and annotators were instructed to record navigation instructions that would be sufficiently descriptive for others to follow the same path. However, this restriction can be relaxed to allow free movement and narration for other purposes. Once the Guide is satisfied with their recording, they are asked to manually transcribe their own voice recording into text. This ensures high quality transcription results.

During the Guide task, in parallel to the annotator's voice recording, PanGEA captures a timestamped record of the annotator's virtual camera movements, which we call a \textit{pose trace}. By default, PanGEA is configured to use Firebase\footnote{\url{https://firebase.google.com}}, saving the Guide's audio recording to a cloud storage bucket, and the transcript, pose trace and other metadata to a cloud database for post processing. Inspired by Localized Narratives \cite{PontTuset_eccv2020}, PanGEA includes a utility to automatically align each Guide's pose trace with the manual transcript of their audio recording. This is achieved by using a Speech to Text service\footnote{\url{https://cloud.google.com/speech-to-text}} to first generate a noisy-but-timestamped automatic transcription. PanGEA then using dynamic time warping to align tokens in the automatic transcript to the manual transcript before propagating timestamps from the automatic to the manual transcription (Figure \ref{fig:pose_trace}). The result is fine-grained synchronization between the transcribed text, the pixels seen, and the actions taken by the Guide. 

\paragraph{Follower Task}
In the Follower task (Figure \ref{fig:ui}, right), Followers begin at a specified starting point in an environment and are asked to follow a Guide's instructions. They observe the environment and navigate as the Guide’s audio plays. Followers can skip forward or backward in the audio recording by clicking on an audio waveform representation of the Guide's recording. This allows them to skip over periods of silence or to listen to part of the audio again. Once the Follower believes they have reached the the end of the path, or they give up, they indicate they are done and the task ends. Note that although the Follower task supports audio instructions, it can be easily adapted to replace the audio instruction with a textual instruction. This was the approach taken by \citet{discriminator21}.

As with the Guide task, the Follower's pose trace is recorded and saved to a cloud database, along with the timestamp of the Guide's audio that the Follower listened to at each moment. This allows the Follower's visual percepts and actions to be accurately aligned with text tokens in the Guide's instructions. Similarity between the annotated (Guide) path and the Follower path is also a natural measure of the joint quality of both the Guide and the Follower annotations. In the experiments for RxR, the path extracted from the Follower's pose trace was also used as additional supervision when training Follower agents, since it represents a step-by-step account of how a human solved the task and the visual inputs they focused on in order to do so \cite{ku2020room}.






\paragraph{Deployment}

PanGEA comes with several demos using a very simplistic environment. To deploy PanGEA for a new large-scale collection effort requires completing 3 main steps:
\begin{compactitem}
\item{Creating a new app in Firebase to initialize the cloud storage and cloud database,}
\item{Setting up an appropriate crowdsourcing platform to serve the PanGEA front-end to a pool of annotators, and}
\item{Setting up the environment to be used, e.g., hosting the images and navigation graphs in a storage bucket in an appropriate format.}
\end{compactitem}
Further details are provided in the PanGEA readme.

\section{Observations and Best Practices}

PanGEA was developed for the collection of the RxR dataset, a 20,000+ hour annotation effort based on Matterport3D indoor scenes. Many of the lessons learned during this collection effort are codified in the PanGEA toolkit. For example, we found that uploading recorded audio at the end of the Guide task was time consuming, and so in the final version of PanGEA the wav file is uploaded in the background while the annotator is busy transcribing their audio. We also found that audio annotations could include long periods of silence, so we provided Follower annotators with an audio waveform visualization and an interface to skip over silence. Some other observations and best practices for reducing annotation times and improving annotation quality are shared in this section.




\paragraph{Annotators Complete Tasks in Creative Ways} 

PanGEA is designed to capture the alignment between annotators' visual percepts, actions and utterances to provide fine-grained spatio-temporal grounding. In initial trials with PanGEA, we found that some annotators -- with the best of intentions -- completely undermined this paradigm. We had envisioned them speaking while moving and looking at the environment; however, in an effort to generate more fluent instructions, some annotators first explored the environment while drafting a navigation instruction separately in a text editor. Then, having finalized the textual instruction, the annotator read it all at the end of the audio recording. 
While this strategy indeed produced high-quality navigation instructions, the instructions were no longer time-aligned to the pose trace. Interestingly, the language used in the instructions also differed. Instructions drafted as text tended to use more connective phrases --- for example, ``turn right \textit{and then} you will see a dining table'' instead of ``turn right... \textit{now} you see a dining table''. We found it challenging to add guardrails in PanGEA that could prevent this behaviour without unduly restricting the freedom of the annotators and the flexibility of the toolkit. Instead, we addressed this issue--successfully--via explicit training.

\paragraph{Annotator Training} To overcome the aforementioned issue and to improve annotation quality in general, for RxR, we conducted an interactive virtual training session with annotators, providing examples of ideal annotations and various failure modes. Annotators were also able to ask questions regarding how to best complete the tasks assigned to them. Although interactive training sessions are not always possible, at minimum we recommend providing annotators with a training video that shows a walk-through of the task and notes common pitfalls to avoid. We provide links to the demo videos for the RxR Guide task\footnote{\url{https://youtu.be/aJkJfB8oI2M}} and Follower task\footnote{\url{https://youtu.be/vcP-oX1t0CU}} (initially called the Tourist task).

\paragraph{Pilot Collections and Learning Periods} Annotating and following navigation instructions in a virtual world is a complex task. We recommend allowing for several small-scale pilot data collections to identify issues with the collection process. This includes having the team creating the dataset perform the tasks using the tool. Secondly, we recommend allowing for a learning period whenever a new annotator is introduced to the task, i.e., planning to discard the first 5--10 annotations produced by a new annotator. We found that rotating annotators between both Guide and Follower tasks early in their experience improved annotation quality because doing so provides much greater awareness of the needs of Followers when completing the Guide tasks.

\paragraph{Data Monitoring Dashboard} We recommend using VLN evaluation metrics such as success rate, navigation error and SPL \cite{anderson2018evaluation} (or similar metrics for alternative tasks) to continually monitor the quality of the collected Guide navigation instructions and Follower paths. By storing the collected annotations in Firebase, it is relatively easy to construct web-based interfaces to monitor these metrics. In the case of RxR, we created a monitoring dashboard that displayed success rates for each annotator pool and also each annotator, with the capability to replay the pose traces from individual Guide and Follower annotations. Annotators were able to see an anonymized view on their progress as it related to others, which helped them assess whether they were performing the task correctly or needed additional changes and perhaps explicit guidance.

\paragraph{Speech versus Writing} In tasks that require a person to perform actions while producing or comprehending language, it is much easier if people are allowed to use speech rather than writing because it allows them to use their hands and eyes fully for performing actions. This has very real consequences for thinking about future data collection efforts. Speech interactions will be essential for any tasks that include time pressure, such as collaborative games where players use language to coordinate. There is also a simple but significant cost advantage: on average, the transcription portion for an RxR Guide annotation took three to four times longer than collecting the speech instruction itself, so either a great deal more instructions could have been collected, or the cost could have been significantly reduced. Speech also encodes intonation and is more likely to elicit interesting dialectal differences. For these and other reasons, we may thus want to encourage more research that works on language grounding tasks that work with speech directly, and provide current best automatic speech recognition output for those who insist on working with text only.

\section{Future Applications}

There are many potential future applications of PanGEA and tools that could be built based on the design decisions discussed above. We are particularly excited about multi-agent problems that collect pose traces from multiple participants as they coordinate via language, such as hide and seek games or tasks where items must be moved from one location to another to satisfy goals or solve puzzles, similar to CerealBar \cite{suhr-etal-2019-executing}.


\bibliography{main}
\bibliographystyle{acl_natbib}

\end{document}